\documentclass[10pt,twocolumn,letterpaper]{article}

\usepackage[pagenumbers]{cvpr}  

\newif\ifdraftmode
\draftmodefalse   

\definecolor{cvprblue}{rgb}{0.21,0.49,0.74}
\usepackage[pagebackref,breaklinks,colorlinks,allcolors=cvprblue]{hyperref}

\def\paperID{0000}
\def\confName{Preprint}
\def\confYear{2026}

\title{Single-Channel EEG-Based Cognitive Load Assessment in Online Learning: A Hybrid Deep Learning Approach}

\author{Rowan Hussein\\
University of Ottawa\\
Ottawa, Ontario, Canada\\
{\tt\small rhuss060@uottawa.ca}
\and
Mohamed Ouf\\
Queen's University\\
Kingston, Ontario, Canada\\
{\tt\small 24blr2@queensu.ca}}

\begin{document}
\maketitle
\begin{abstract}
Monitoring cognitive load during online learning could help instructors identify content that learners find difficult, but remote settings remove the visual cues that support this judgement in a classroom. We study whether a single-channel, consumer-grade EEG device (the NeuroSky MindWave Mobile 2) can distinguish easy from difficult educational-video content, using the publicly available dataset of Wang et al.~\cite{wang2013using} (ten learners, one excluded for excessive noise, leaving nine). We implement a hybrid CNN+LSTM+Attention model that combines the raw waveform with band-power features. In a within-subject setting, the model reaches up to $78.5\%$ accuracy, compared with $55\%$ for conventional feature-based classifiers; regularization (dropout and $L2$) closes the large gap between training and validation accuracy that we observe without it, keeping validation accuracy stable at roughly $68$--$73\%$. We are deliberately cautious about these numbers: with only nine subjects, within-subject evaluation is optimistic, and we argue that \emph{subject-independent} evaluation---in which no learner appears in both training and test data---should be the standard for this task. To that end we release a reproducible evaluation pipeline. We frame the work as a feasibility study rather than a deployable system, and pair it with an open, notebook-based tool that records EEG, runs inference, and visualizes estimated cognitive load as a heatmap over the video timeline to help educators locate potentially challenging segments.
\end{abstract}
    
    \section{Introduction}
\label{sec:intro}

The rapid transition to online education following the COVID-19 pandemic highlighted a number of critical challenges in maintaining effective instruction and meaningful student engagement. Traditional classroom dynamics, where instructors can visually assess comprehension and adjust teaching strategies in real-time, were largely lost in virtual settings. As a result, many online learners experienced reduced motivation, diminished participation, and less interactive communication with peers and educators \cite{Hodges2020, Bao2020}. The absence of face-to-face feedback mechanisms and the isolating nature of remote learning environments hinder instructors' ability to gauge understanding and provide help. Consequently, these conditions may negatively impact learning outcomes and overall educational effectiveness.

Electroencephalography (EEG) offers a promising avenue to address these issues by enabling the continuous monitoring of cognitive load during online learning sessions. By capturing the brain’s electrical activity, EEG can detect subtle shifts in mental effort and engagement as learners interact with educational content \cite{Antonenko2010, Anderson2011}. Real-time EEG-based feedback allows educators to identify students who struggle with specific topics, encouraging timely interventions such as clarifying explanations or adjusting teaching materials. Moreover, when a significant portion of the class shows elevated cognitive load at the same point in the instructional material, it may indicate that the content is inherently challenging, prompting educators to reevaluate their instructional design and delivery.
\begin{figure}[!t]
\centering
\includegraphics[width=0.4
\columnwidth]{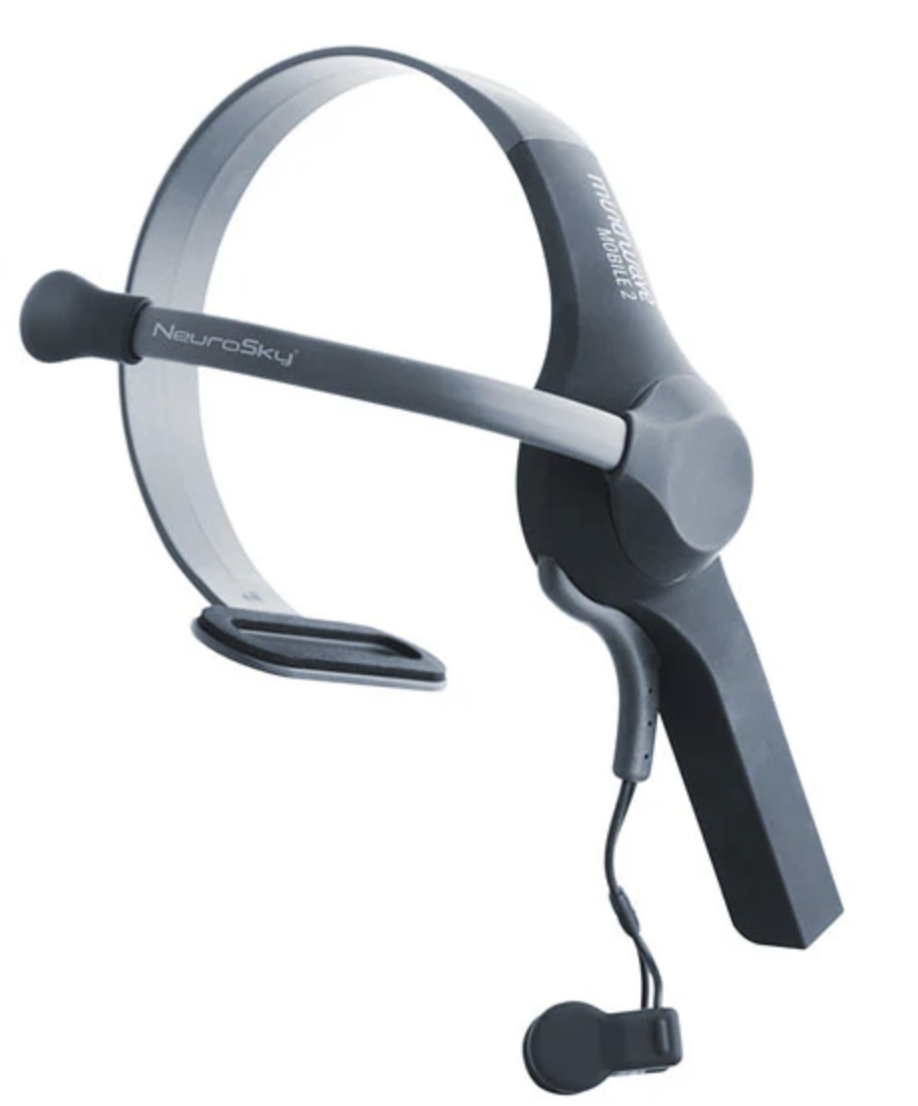}
\caption{MindWave Mobile 2.}
\label{fig:sky}
\end{figure}

Beyond the measurement of basic mental states, EEG has also been used extensively in medical and biomedical applications, including the analysis of mental workload and fatigue \cite{Kather2020}, the diagnosis of neurological conditions \cite{Lotte2007}, and the rehabilitation of disorders affecting the central nervous system \cite{Daly2008}. In educational contexts, EEG-based analyses have proven valuable for understanding learners' cognitive processes, emotional states, and engagement patterns \cite{Zhang2020}. These insights are especially useful given recent advancements in consumer-grade EEG devices. These systems simplify data acquisition by eliminating the need for complex multi-channel setups, conductive gels, and extensive calibration procedures. For instance, single-channel EEG headsets, such as the MindWave Mobile 2 (Figure~\ref{fig:sky}), are similar in size to conventional headphones and require only minimal preparation, thus lowering barriers to adoption in real-world educational environments \cite{Naseer2015, NeuroskyWebsite}.

Despite such technological and methodological advancements, accurately measuring cognitive load from EEG signals remains a challenging task, particularly when working with small datasets. State-of-the-art approaches often rely on multi-channel EEG recordings and leverage sophisticated machine learning models—such as Convolutional Neural Networks (CNNs), Long Short-Term Memory (LSTM) networks, and attention-based architectures—to classify cognitive load states \cite{Zhang2020, Li2021, Sarkar2022}. While these methods have demonstrated commendable performance on larger datasets, smaller datasets are prone to overfitting and reduced model generalizability \cite{Rasheed2021}. Strategies like regularization, cross-validation, and careful preprocessing are required to mitigate these limitations and ensure reliable outcomes.

In this study, we present a single-channel EEG-based approach to estimating cognitive load while learners watch educational videos. Our hybrid CNN+LSTM+Attention model, which combines the raw waveform with band-power features, reaches up to $78.5\%$ accuracy in distinguishing easy from difficult content in a within-subject setting, compared with $55\%$ for conventional feature-based classifiers. We treat this result cautiously: with only nine subjects, within-subject evaluation is optimistic, and we argue that \emph{subject-independent} evaluation---in which no learner contributes data to both training and testing---should be the standard for judging whether such a system generalizes to new learners. To support this, we release a reproducible evaluation pipeline. We build the model into a processing pipeline that integrates data acquisition from a MindWave Mobile~2 headset, signal preprocessing, feature extraction, and inference; by visualizing the resulting estimates as a heatmap over the video timeline, it gives educators intuitive feedback for locating segments that may warrant intervention.

\noindent\textbf{Contributions:} This work makes three contributions:

\textbf{A feasibility study on consumer single-channel EEG.} We show that a hybrid CNN+LSTM+Attention model reaches up to $78.5\%$ within-subject accuracy on the confusion-detection task, substantially above conventional feature-based classifiers, and we characterize its overfitting behaviour and the effect of regularization.

\textbf{A methodological caution and open evaluation pipeline.} We argue that subject-independent evaluation should be the default for this task, and we release code that implements it---leave-one-subject-out cross-validation, an explicit chance baseline, a label-permutation significance test, and a branch ablation---so that generalization to unseen learners can be assessed reproducibly.

\textbf{An open visualization tool.} We provide a simple, notebook-based tool that, once an EEG session is recorded, processes the data and displays a cognitive-load heatmap over the video timeline.

\noindent\textbf{Organization:} Following this introduction, we review related work on EEG-based cognitive load assessment and machine learning techniques. We then present our methodology, including data collection, preprocessing, and model development. Next, we discuss the results and their implications. Finally, we conclude with a summary of our findings and directions for future research.

\section{Literature Review}
\label{sec:lit_review}

Electroencephalography (EEG) has been extensively employed to measure cognitive load and emotional states in both educational and human-computer interaction (HCI) contexts. Traditional methodologies often utilize multi-channel EEG systems, allowing for the extraction of rich neural patterns associated with varying cognitive engagement and emotional responses \cite{Antonenko2010, Anderson2011}. For instance, Zarjam et al. \cite{Zarjam2011} identified specific spectral features from EEG signals that effectively discriminate between different cognitive load levels during reading tasks. Similarly, Anderson et al. \cite{Anderson2011} conducted a user study examining the cognitive load imposed by distinct visualization techniques (e.g., graphical versus textual representations). Their findings suggest that EEG-based assessments can inform the selection and refinement of visualization methods, potentially guiding the creation of more effective educational materials. Antonenko et al. \cite{Antonenko2010} further demonstrated the utility of continuous EEG measurements to capture subtle fluctuations in cognitive load, providing valuable insights into how particular instructional interventions influence cognitive processing.

Recent advancements in machine learning (ML) and deep learning (DL) have significantly contributed to the analysis and classification of EEG data for cognitive load assessment. Numerous approaches, including Convolutional Neural Networks (CNNs), Long Short-Term Memory (LSTM) networks, and attention-based architectures, have been explored to improve the accuracy and robustness of cognitive load classification \cite{Zhang2020, Sarkar2022, Li2021}. For example, Zhang et al. \cite{Zhang2020} introduced a deep attention-based LSTM framework that effectively classified hand movements from EEG signals, outperforming traditional models. Likewise, Li et al. \cite{Li2021} employed a multilevel multiscale CNN for motor imagery EEG, achieving high classification accuracies by capturing time-frequency-space features. One recurring challenge is the limited size of available EEG datasets, which can lead to overfitting and reduced generalizability \cite{Rasheed2021, Luo2018}, and---as we emphasize in this paper---to over-optimistic accuracy estimates when the evaluation split allows data from the same subject to appear in both training and test sets. Some studies address the sample-size limitation with careful preprocessing pipelines and regularization \cite{Sarkar2022, Li2021}. Another promising development is the emergence of consumer-grade EEG devices---such as the NeuroSky MindWave, Muse, and Emotiv Insight~\cite{NeuroskyWebsite, MuseWebsite, EmotivWebsite}---which simplify data acquisition and circumvent the complexities inherent in multi-channel EEG systems \cite{Naseer2015}.
These portable and user-friendly solutions enable broader experimentation. Machine learning has proven effective across a range of human-centered sensing tasks---for example, computer-vision-based injury prevention and activity classification in physical fitness~\cite{ouf2024visionpf, hussein2022lstm}---as well as in software-engineering analytics~\cite{ouf2025reverse, ouf2026empirical, ouf2026dogood, ouf2026same} and communications and maritime monitoring systems~\cite{marshall2024ris, soliman2024bibliometric}; we bring this data-driven perspective to EEG-based cognitive-load estimation. Building on these trends, our study asks how well a single-channel EEG device can estimate cognitive load when the model is evaluated on unseen learners, and offers an accessible, reproducible pipeline for online-learning settings.

\section{Methodology}
\label{sec:meth}

This section outlines our approach to collecting, processing, and analyzing the EEG dataset, as well as the development and refinement of our machine learning models. We describe the data source, preprocessing procedures, feature extraction methods, model architectures, and the steps taken to mitigate overfitting. Throughout, we highlight our iterative process of trial and error, guiding the selection of techniques aimed at improving classification performance under challenging data constraints.

\subsection{Data Collection}

\subsubsection{Data Source}
We employed a dataset originally presented by Wang et al.~\cite{wang2013using}, where EEG signals were recorded from ten college students as they watched educational video segments of varying difficulty. A pool of 20 videos (approximately two minutes each) was used, with half categorized as “easy” and the other half as “hard.” To introduce potential confusion, two-minute clips were extracted from the middle of a topic. Each participant watched 10 of these videos (5 easy, 5 hard), giving a total of 100 trials. After viewing each video, participants provided a self-reported confusion rating on a 7-point Likert scale, later normalized into a binary label (confused vs. not confused). In addition to these \textit{user-defined labels}, \textit{predefined labels} based on the intended difficulty of the videos were also available.

Participants, aged 24–31 and from diverse backgrounds, had normal or corrected-to-normal vision and no reported neurological conditions. Informed consent was obtained, and ethical approval was granted by the relevant institutional review board. After excluding one participant (Subject~\#6) whose signal was dominated by noise (Section~\ref{sec:cleaning}), our analysis uses the remaining nine subjects (90 trials); we report counts consistently on this post-exclusion set.

\subsubsection{Data Acquisition Device}
The EEG signals were acquired using a single-channel NeuroSky MindWave Mobile~2 headset (Figure~\ref{fig:sky}), with the primary sensor positioned near the FP1 location and the reference electrode at the ear clip. The device streams the raw single-channel waveform at approximately 512~Hz over Bluetooth and also reports derived metrics (e.g., band-power values) computed on-board via the Fast Fourier Transform (FFT) and proprietary algorithms. This plug-and-play configuration eliminates the need for complex multi-channel setups or gel-based electrodes, making the device accessible and suitable for the educational context.

\subsection{Data Processing and Preprocessing}

\subsubsection{Data Cleaning}
\label{sec:cleaning}
Initial variance analysis indicated that several device-generated features (e.g., the proprietary attention and meditation indices and certain band powers) offered limited discriminatory value due to their low variance. Consequently, we based the CNN branch on the raw EEG signal and reserved the band-power features for the recurrent branch (Section~\ref{sec:features}).

\begin{figure}[!t]
\centering
\includegraphics[width=0.9\columnwidth]{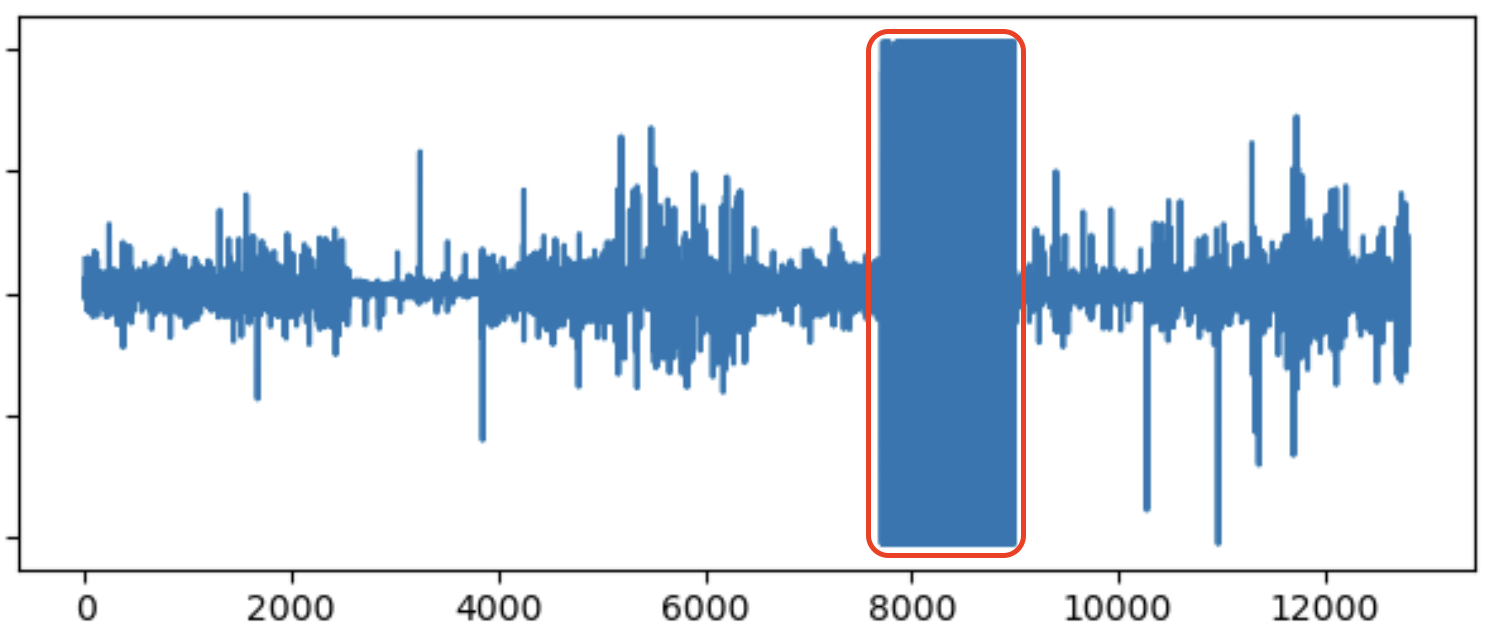}
\caption{Raw single-channel EEG for the excluded Subject~\#6, showing the large-amplitude noise that motivated its removal.}
\label{fig:user6}
\end{figure}

A manual inspection revealed that one participant's data (Subject~\#6, Figure~\ref{fig:user6}) were predominantly noisy and thus excluded. Other than this removal, the dataset was relatively complete and required minimal handling of missing values. The final dataset associates each trial with its raw EEG time series, band-power sequence, and the corresponding confusion label.

\subsubsection{Feature Extraction}
\label{sec:features}
EEG signals are non-stationary, so it is important to capture both time- and frequency-domain characteristics. We segment each trial into short windows and derive two complementary representations that feed the two model branches. The first is the (normalized) raw waveform itself, provided as a fixed-length window to the convolutional branch; we treat the window length as a hyperparameter and study its effect in Section~\ref{sec:window}. The second is a sequence of relative band-power vectors: for each of a small number of consecutive sub-windows we compute the relative power in the \textit{delta} (0.5–4~Hz), \textit{theta} (4–8~Hz), \textit{alpha} (8–12~Hz), and \textit{beta} (12–30~Hz) bands, yielding a length-$10$ sequence of $4$-dimensional vectors for the recurrent branch. We additionally computed a set of time-domain descriptors---\textit{mean}, \textit{variance}, \textit{skewness}, \textit{kurtosis}, \textit{zero-crossings}, \textit{absolute area under the curve}, and \textit{peak-to-peak distance}---which we use for the simple feature-based baseline (Section~\ref{sec:eval}).

\begin{table}[!t]
\centering
\caption{Equations for Selected Features and Regularization}
\begin{tabular}{@{}ll@{}}
\toprule
\textbf{Feature/Term} & \textbf{Equation} \\ \midrule
Mean $(\mu)$ & $\displaystyle \mu = \frac{1}{N}\sum_{i=1}^{N} x_i$ \\[8pt]

Variance $(\sigma^2)$ & $\displaystyle \sigma^2 = \frac{1}{N}\sum_{i=1}^{N}(x_i - \mu)^2$ \\[8pt]

Skewness & $\displaystyle \text{Skew} = \frac{1}{N}\sum_{i=1}^{N}\left(\frac{x_i-\mu}{\sigma}\right)^3$ \\[8pt]

Kurtosis & $\displaystyle \text{Kurt} = \frac{1}{N}\sum_{i=1}^{N}\left(\frac{x_i-\mu}{\sigma}\right)^4 - 3$ \\[8pt]

Zero-Crossings & Count of sign changes in $\{x_i\}_{i=1}^N$ \\[8pt]

Absolute Area & $\displaystyle A = \sum_{i=1}^{N} |x_i|$ \\[8pt]

Peak-to-Peak & $\displaystyle p2p = x_{\max} - x_{\min}$ \\[8pt]

Relative Band Power & $\displaystyle P_{\text{band}} = \frac{\sum_{f \in \text{band}} |X(f)|^2}{\sum_{f} |X(f)|^2}$ \\[8pt]

L2 Regularization & $\displaystyle \text{Loss}_{L2} = \text{Loss}_{\text{original}} + \lambda \sum_{j} w_j^2$ \\

\bottomrule
\end{tabular}
\label{tab:feature_eq}
\end{table}

Table~\ref{tab:feature_eq} summarizes the mathematical definitions of the key features and the L2 regularization term.

\subsubsection{Data Segmentation for Sequence Modeling}
To facilitate sequence modeling, we used a moving-window approach and tested several raw-window lengths ($25$, $50$, and $100$ samples, i.e., roughly $0.05$--$0.20$~s at the device's $\sim$512~Hz raw rate) to identify the temporal granularity that best captured cognitive-load variations while balancing temporal resolution against the number of usable segments per trial. Because these windows are short relative to a two-minute trial, this study is a within-subject sensitivity analysis (Section~\ref{sec:window}), not our headline generalization result.

\subsection{Model Development}

\subsubsection{Baseline Models}
We initially evaluated conventional machine learning classifiers, including Decision Trees, Random Forests, XGBoost, Support Vector Machines, and Logistic Regression. These models served as baselines, achieving modest accuracies (roughly 25–55\%) when predicting confusion states. While these results were modest, they provided a benchmark and underscored the inherent difficulty of the classification task.

\subsubsection{Advanced Models (LSTM and Attention)}
To better exploit temporal dependencies in the EEG signals, we tried Long Short-Term Memory (LSTM) networks, building on our prior experience applying LSTMs to sequence classification of activity videos~\cite{hussein2022lstm}. LSTMs capture long-term temporal structures, which can be valuable for identifying subtle patterns in cognitive load. Additionally, we integrated an attention mechanism to focus on the most informative time steps, similar to that of Zhang et al.~\cite{Zhang2020}. While this approach improved representational capacity, the small dataset size led to persistent overfitting challenges.

\subsubsection{Proposed CNN+LSTM+Attention Architecture}
\begin{figure}[!t]
\centering
\includegraphics[width=0.9\columnwidth]{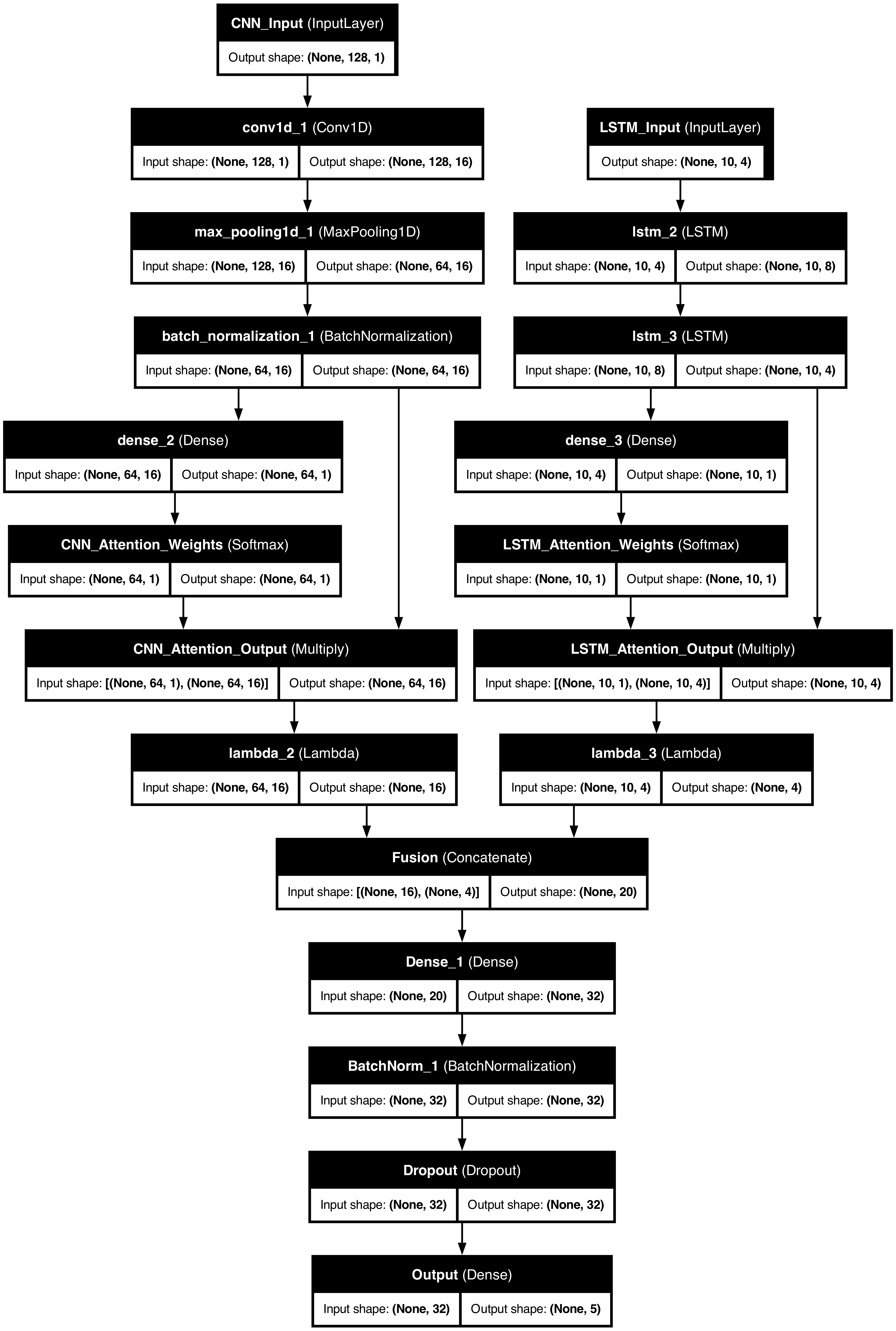}
\caption{Schematic of the proposed dual-input CNN+LSTM+Attention architecture (convolutional branch, recurrent branch, per-branch attention, and fusion). The evaluated model uses the window length selected in Section~\ref{sec:window} and a single-sigmoid binary output (see footnote).}
\label{fig:model}
\end{figure}

Building on these trials, we developed a dual-input hybrid architecture (Figure~\ref{fig:model}). The convolutional branch takes the raw window (Section~\ref{sec:window}) and applies 1D convolution, max-pooling, and batch normalization to extract local temporal patterns; the recurrent branch takes the length-$10$ sequence of $4$-dimensional band-power vectors and applies stacked LSTMs to model longer-range dependencies. Each branch has its own additive attention layer (a softmax over time steps) that reweights and pools its features. The two pooled representations are concatenated and passed through a fully connected layer with batch normalization, dropout, and $L2$ regularization. Because the task is binary (confused vs.\ not confused), the output is a \emph{single sigmoid unit} trained with binary cross-entropy.\footnote{Figure~\ref{fig:model} illustrates the dual-branch structure; the specific input and output dimensions it shows correspond to an earlier configuration (a longer input window and a multi-output head), whereas the model we evaluate uses the window length selected in Section~\ref{sec:window} and the single-sigmoid binary head.}

\subsubsection{Overfitting Mitigation}
Given the small sample, we combined \textbf{dropout} (randomly dropping units in the fully connected layer), \textbf{$L2$ regularization} (penalizing large weights), and \textbf{early stopping}. No data augmentation was used at this stage, though synthetic data generation is a natural direction for future work.

\subsection{Evaluation Protocol}
\label{sec:eval}
The results reported in this paper use \textbf{within-subject} evaluation, with accuracy computed on held-out windows and cross-validated over the segmented data. This is the standard setting in much small-sample EEG work and is what we use to compare models and window sizes.

We stress, however, that within-subject evaluation is \emph{optimistic} for our purpose. With only nine subjects and many short windows per trial, windows from the same subject can appear in both training and test sets, so a model may exploit subject- and trial-specific idiosyncrasies rather than generalizable correlates of cognitive load---a well-documented pitfall. For a system meant to help new learners, the appropriate protocol is \textbf{subject-independent leave-one-subject-out (LOSO) cross-validation}, in which each subject is held out entirely for testing, paired with a majority-class chance baseline, a label-permutation significance test, and a branch ablation under identical folds. We do not report LOSO numbers as a settled result here; instead we release code that runs this full protocol (Section~\ref{sec:generalization}) so that generalization can be measured reproducibly. This distinction---and the caution it implies---is a central message of the paper.

\subsection{Implementation Details}
All experiments were conducted in \textit{Python} on a \textit{MacBook Pro (M2 Pro)}. Models were built in \textit{TensorFlow}/\textit{Keras}; \textit{scikit-learn} provided the baselines, cross-validation splitters, and metrics; and \textit{NumPy}/\textit{Pandas} handled data manipulation, with \textit{Matplotlib}/\textit{Seaborn} for visualization. All random seeds (\textit{NumPy} and \textit{TensorFlow}) were fixed and are reported in the released code to support reproducibility.

\section{Results}

We first report conventional baselines, then the performance of the proposed hybrid model, its overfitting behaviour, and a within-subject window-size analysis. We close with a qualitative visualization and a discussion of why subject-independent evaluation is the essential next step.

\subsection{Baseline Model Performance}

We first evaluated a range of conventional classifiers---Random Forest, XGBoost, SVM, Logistic Regression, Decision Tree, Gradient Boosting, AdaBoost, and KNN---on hand-crafted features, using both the predefined (content-difficulty) and user-defined (self-reported confusion) labels. As shown in Table~\ref{tab:baseline_results}, performance is modest. Several accuracies fall at or below the $50\%$ chance level of this balanced binary task; this reflects the very small per-fold test sets and the difficulty of decoding confusion from a single channel, and it motivates both the sequence-based model below and our caution about single split-based numbers. The best user-defined-label accuracy is around $55\%$ (Decision Tree, KNN, Logistic Regression).

\begin{table}[!t]
\centering
\caption{Performance of baseline classifiers on predefined and user-defined labels. These conventional models establish the difficulty of the task.}
\resizebox{\columnwidth}{!}{%
\begin{tabular}{lcccccccc}
\toprule
 & \multicolumn{4}{c}{\textbf{Predefined Label}} & \multicolumn{4}{c}{\textbf{User-Defined Label}} \\ \cmidrule(lr){2-5}\cmidrule(lr){6-9}
\textbf{Model} & \textbf{Acc} & \textbf{Prec} & \textbf{Rec} & \textbf{F1} & \textbf{Acc} & \textbf{Prec} & \textbf{Rec} & \textbf{F1}\\
\midrule
Random Forest       & 0.35 & 0.6250 & 0.35 & 0.3321 & 0.35 & 0.4344 & 0.35 & 0.3051 \\
XGBoost             & 0.45 & 0.6250 & 0.45 & 0.4795 & 0.50 & 0.5303 & 0.50 & 0.5104 \\
SVM                 & 0.25 & 0.0625 & 0.25 & 0.1000 & 0.35 & 0.1225 & 0.35 & 0.1815 \\
Logistic Reg.       & 0.50 & 0.6515 & 0.50 & 0.5313 & 0.55 & 0.5345 & 0.55 & 0.5410 \\
Decision Tree       & 0.55 & 0.7396 & 0.55 & 0.5742 & 0.55 & 0.6917 & 0.55 & 0.5421 \\
Gradient Boosting   & 0.40 & 0.6667 & 0.40 & 0.4000 & 0.50 & 0.5303 & 0.50 & 0.5104 \\
AdaBoost            & 0.35 & 0.6250 & 0.35 & 0.3321 & 0.50 & 0.5606 & 0.50 & 0.5101 \\
KNN                 & 0.35 & 0.5536 & 0.35 & 0.3647 & 0.55 & 0.4938 & 0.55 & 0.5119 \\
\bottomrule
\end{tabular}}
\label{tab:baseline_results}
\end{table}

\subsection{Proposed Hybrid Model}
\label{sec:main}

Moving beyond hand-crafted features, the proposed dual-input CNN+LSTM+Attention model leverages both the raw waveform and the band-power sequence. In a within-subject setting it reaches up to $78.5\%$ accuracy (Section~\ref{sec:window}), a clear improvement over the conventional classifiers in Table~\ref{tab:baseline_results}, whose best result is $55\%$. This supports the intuition that jointly modelling local waveform structure (via the convolutional branch) and longer-range spectral dynamics (via the recurrent branch), with attention over time steps, captures information that static feature summaries miss. We emphasize, however, that these are within-subject numbers on a nine-subject dataset and should be read as evidence of feasibility rather than as a generalization estimate; Section~\ref{sec:generalization} returns to this point.

\subsection{Overfitting Analysis and Mitigation}

\begin{figure}[!t]
\centering
\includegraphics[width=\columnwidth]{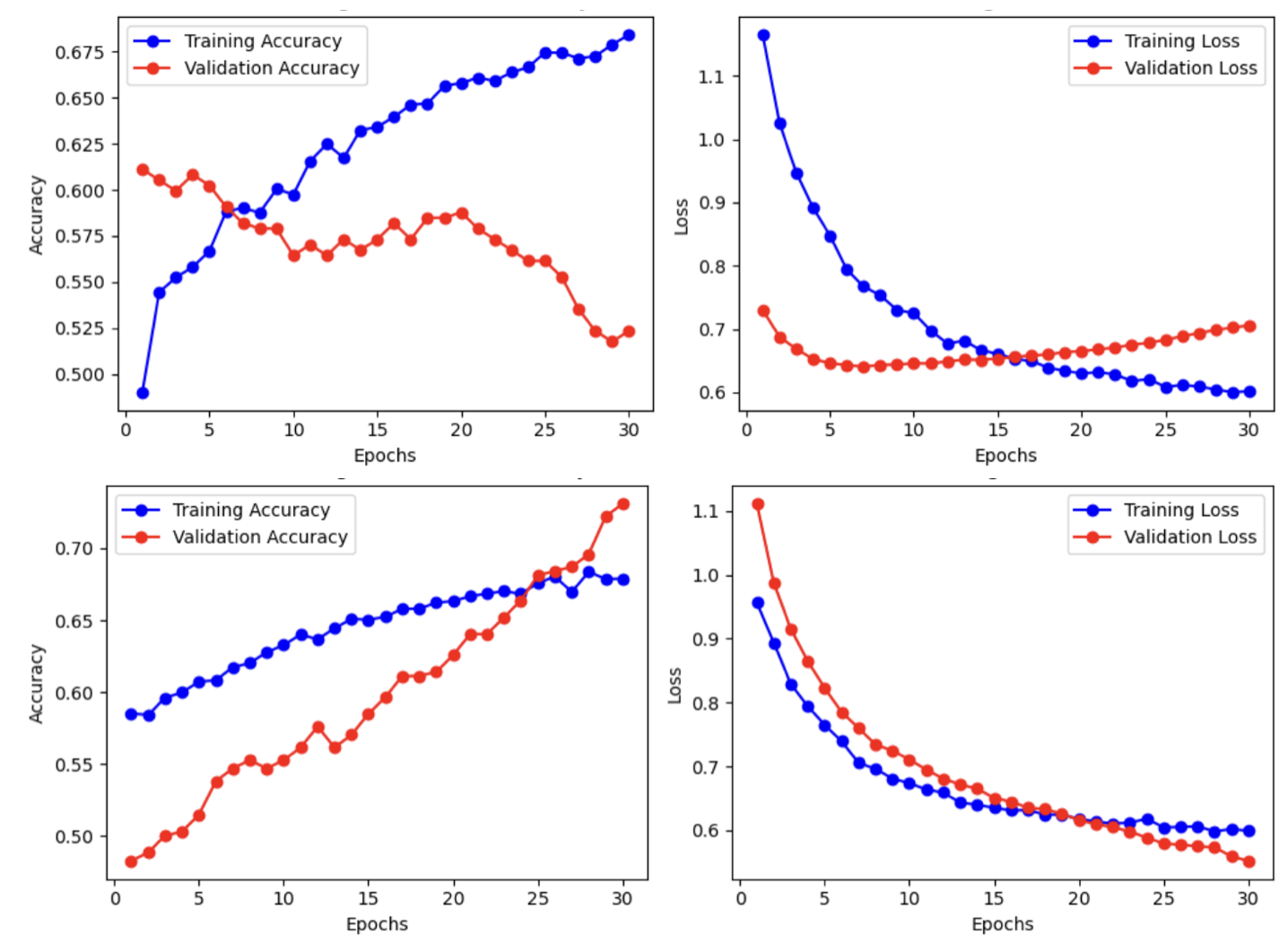}
\caption{Training and validation accuracy/loss for the proposed model before (top) and after (bottom) applying dropout and $L2$ regularization. After regularization, validation accuracy stabilizes around $0.68$--$0.73$ and the train--validation gap shrinks.}
\label{fig:overfitting_mitigation}
\end{figure}

Overfitting was a recurring challenge: early models reached high training accuracy but generalized poorly. Dropout, $L2$ regularization, and early stopping narrowed this gap. Figure~\ref{fig:overfitting_mitigation} shows the effect directly. Before regularization (top), validation accuracy peaks near $0.61$ and then declines while training accuracy keeps climbing---the classic overfitting signature. After regularization (bottom), the two curves track together and plateau around $0.68$--$0.73$. This behaviour is exactly what one expects on a small dataset, and it is why we report regularized performance and treat the raw peak accuracies with caution.

\subsection{Effect of Window Size}
\label{sec:window}

We examined three raw-window sizes ($25$, $50$, and $100$ samples). As shown in Table~\ref{tab:window_size_results}, a window of $50$ samples gave the best accuracy ($78.5\%$), with $25$ and $100$ samples at $75.0\%$ and $76.0\%$. The drop at $100$ samples likely stems from fewer usable segments per trial and the resulting padding, which reduces the effective amount of training data. We therefore use a $50$-sample window in the proposed model.

\begin{table}[!t]
\centering
\caption{Within-subject classification accuracy of the proposed model for different raw-window sizes.}
\begin{tabular}{lc}
\toprule
\textbf{Window Size (samples)} & \textbf{Accuracy (\%)} \\
\midrule
25 & 75.0 \\
50 & 78.5 \\
100 & 76.0 \\
\bottomrule
\end{tabular}
\label{tab:window_size_results}
\end{table}

\subsection{Qualitative Visualization Over Time}

\begin{figure}[!t]
\centering
\includegraphics[width=0.8\columnwidth]{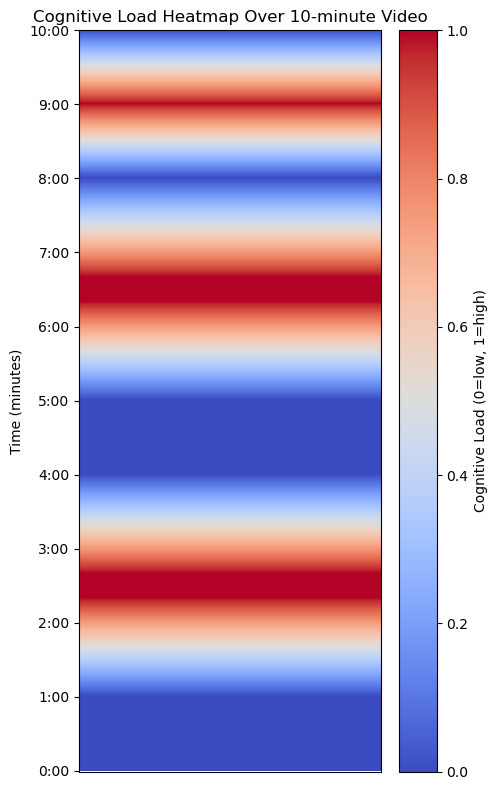}
\caption{Illustrative cognitive-load heatmap produced by the visualization pipeline over a concatenated video, intended to show the interface rather than to serve as a quantitative result.}
\label{fig:cognitive_load_heatmap}
\end{figure}

Beyond aggregate metrics, the pipeline can render cognitive-load estimates over the course of a video. As a qualitative demonstration, we concatenated several segments into a single longer clip containing three challenging (hard) segments interspersed among easier content, ran inference, and applied a median filter to the model's output to suppress transient artifacts. Figure~\ref{fig:cognitive_load_heatmap} shows the resulting heatmap over the video timeline. We present this as an illustration of the interface rather than a quantitative result; systematically measuring how well predicted high-load regions align with known difficult segments is left to future work.

\subsection{Toward Subject-Independent Evaluation}
\label{sec:generalization}

The results above use within-subject evaluation, which is standard in much small-sample EEG work but is optimistic: because short windows from the same subject can appear in both training and test sets, a model can exploit subject-specific signal characteristics rather than generalizable correlates of cognitive load. For a system intended to help \emph{new} learners, the more honest question is how well it transfers to subjects it has never seen. The appropriate protocol is subject-independent leave-one-subject-out (LOSO) cross-validation, paired with a majority-class chance baseline and a label-permutation significance test. We expect subject-independent accuracy to be lower than the within-subject figures reported here---consistent with the validation plateau of $0.68$--$0.73$ in Figure~\ref{fig:overfitting_mitigation}---and we regard establishing that number as the essential next step rather than a settled result. To make this straightforward and reproducible, we release an evaluation script that implements LOSO cross-validation, the chance baseline, the permutation test, and a branch ablation (CNN-only, LSTM-only, and the full model) under identical folds.

\section{Conclusion and Future Work}
\label{sec:conclusion}

This study examined whether cognitive load can be assessed from a single-channel, consumer-grade EEG device while learners watch educational videos. A hybrid CNN+LSTM+Attention model that combines the raw waveform with band-power features reached up to $78.5\%$ within-subject accuracy on a nine-subject dataset, clearly above conventional feature-based classifiers ($55\%$), while regularization kept validation accuracy stable around $68$--$73\%$. Alongside the model, we provide an open, notebook-based pipeline that records EEG, runs inference, and visualizes estimated cognitive load as a heatmap over the video timeline.

\noindent\textbf{Limitations.} The dataset is small (nine subjects after exclusion), the signal is a single channel from a consumer device, and the labels are coarse. Our headline numbers are within-subject and therefore optimistic; we accordingly frame this as a reproducible feasibility study rather than a deployable system, and we avoid claims of clinical or classroom reliability.

\noindent\textbf{Future Work.} The most important next step is a full subject-independent (leave-one-subject-out) evaluation with a chance baseline and significance testing, for which we release code (Section~\ref{sec:generalization}). Beyond that, promising directions include multi-channel or additional-sensor inputs to capture richer neural activity, larger and more diverse datasets, domain adaptation and data augmentation to improve robustness in data-limited regimes, and interpretability methods to relate features to cognitive load. Integrating the pipeline into online-learning platforms could then support real-time, personalized instructional adjustments.

{
    \small
    \bibliographystyle{ieeenat_fullname}
    \bibliography{main}
}


\end{document}